# Superpixel based Class-Semantic Texton Occurrences for Natural Roadside Vegetation Segmentation

Ligang Zhang* and Brijesh Verma

*School of Engineering and Technology, Central Queensland University, Brisbane, Australia*
{l.zhang, b.verma}@cqu.edu.au

**Abstract:** Vegetation segmentation from roadside data is a field that has received relatively little attention in present studies, but can be of great potentials in a wide range of real-world applications, such as road safety assessment and vegetation condition monitoring. In this paper, we present a novel approach that generates class-semantic color-texture textons and aggregates superpixel based texton occurrences for vegetation segmentation in natural roadside images. Pixel-level class-semantic textons are first learnt by generating two individual sets of bag-of-word visual dictionaries from color and filter-bank texture features separately for each object class using manually cropped training data. For a testing image, it is first oversegmented into a set of homogeneous superpixels. The color and texture features of all pixels in each superpixel are extracted and further mapped to one of the learnt textons using the nearest distance metric, resulting in a color and a texture texton occurrence matrix. The color and texture texton occurrences are aggregated using a linear mixing method over each superpixel and the segmentation is finally achieved using a simple yet effective majority voting strategy. Evaluations on two public image datasets from videos collected by the Department of Transport and Main Roads (DTMR), Queensland, Australia, and a public roadside grass dataset show high accuracy of the proposed approach. We also demonstrate the effectiveness of the approach for vegetation segmentation in real-world scenarios.

**Keywords:** Vegetation segmentation, texton, superpixel, classification algorithm, object recognition

## 1. Introduction

Reliably classifying the types of vegetation present on natural roadside data is important for many real-world applications such as vehicle moving navigation, road risk assessment and identification, and vegetation growth monitoring and management. It is a daunting challenge to develop techniques for robust segmentation in a natural condition due to the presence of substantial variations in the scene content, environmental conditions or capturing settings etc. A high degree of unstructured, dynamic or even unpredictable configuration of vegetation may present in the forms of various intensities, color, texture, shapes, and geometrical locations. The environmental conditions may also have dramatic variations depending on factors such as the daytime, season, location and weather conditions, leading to complicated environmental effects, such as shadows of objects, sunlight reflectance, and a dim or bright lighting condition etc. The data capturing settings may also impact the quality of the data, and the scenes may be overexposed, underexposed, blurred, or of low-resolution or different camera viewpoints, etc.

---

*Corresponding author. Telephone & Fax: +(61) 0732951162.

Although vegetation has long been researched in the fields of remote sensing [1], agriculture, and ecosystem using satellite and aerial data, vegetation segmentation on roadside data is a relatively less investigated field. Compared with satellite and aerial data, ground data has the advantage of requiring much less costs, enabling easy operations, and supporting location-specific precise analysis. Previous work on ground data can be broadly grouped into two categories – visible approaches which analyze the visual characteristics of vegetation in the visible spectrum and seek to utilize color, texture, shape, geometry, and structure features from the visible spectrum to distinguish vegetation from other objects; and invisible approaches which focus on the use of the spectral properties of chlorophyll-rich vegetation and their reflectance characteristics in the invisible spectrum, particularly vegetation indices (VIs). It remains a challenging task to select suitable visible features and design reliable VIs for natural conditions.

Recently, textons have demonstrated as a compact and effective representation of visual characteristics for object categorization. However, existing texton based approaches are primarily based on filter bank responses, which represent only texture features, and often aggregate textons into a histogram profile for each image, which requires a responsibly large image size to avoid sparse histogram bins. Some studies [2] have explored the use of textons for road scene understanding, but little attention has been given specifically to roadside vegetation segmentation [3].

This paper proposes a novel color-texture texton based approach for natural roadside vegetation segmentation. It generates class-semantic textons of color and filter-bank based texture responses from the training data, which represent the intrinsic features learnt for each class. Features in a test image are quantized into one of the learnt textons. The segmentation is then achieved by aggregating color-texture texton occurrences over each superpixel using a simple yet effective majority voting strategy. We demonstrate promising performance of the approach in segmenting vegetation on public natural datasets.

The rest of the paper is organized as follows. Section 2 presents an overview of the related work on vegetation segmentation and the major contributions of this paper. Section 3 introduces the proposed approach. The experiments and result analysis are presented in Section 4. A discussion is presented in Section 5. Finally, the paper is concluded in Section 6.

## 2. Related Work and Contributions

This section reviews prior work on vegetation segmentation using visible features and discusses existing approaches for generic object segmentation that can be potentially used for roadside vegetation analysis. In addition, we also briefly describe related texton based approaches and highlight our contributions.

*2.1 Vegetation Segmentation Approach*

Vegetation segmentation approaches distinguish vegetation from other objects, such as soil, tree, car and road, by exploring their discriminative characteristics in the visible or invisible spectrum. This paper focuses on only visible approaches, which generally utilize the color, shape, texture, geometry and structure features. A major benefit of visible features is that they retain high consistency with human visual perception. Color is one of the dominant resources that the human eyes depend on in the perception and discrimination of different objects, and it is one of most popular features in existing research on vegetation segmentation, which mainly focuses on investigating the suitability of different types of color spaces, including CIELab [4], YUV [5], HSV [6], and RGB [7] etc. Although vegetation is widely known as being characterized primarily by a green or orange color, it is still a challenge to find a suitable color representation of vegetation in complex natural conditions. For instance, vegetation color is theoretically believed to be green in the HSV space under most environment conditions. However, it may not be the case in scenes containing sky and with varying lighting conditions, such as the presence of shadow, shining, under- and over-exposure effects. Another popular type of visible features is texture, which is often represented by performing wavelet filters, such as Gabor filters [8] and Continuous Wavelet Transform (CWT) [6], extracting pixel intensity distributions, such as pixel intensity differences (PIDs) [4], [5] and variations in a neighborhood [9], [10], or generating spatial statistic measures [10], entropy [7], or statistical features over superpixels [11].

Table 1 lists typical visible approaches for vegetation segmentation in existing studies. One of the early studies on vegetation segmentation in outdoor images was presented in [12], which employed a self-organising map (SOM) for object segmentation and then extracted color, texture, shape, size, centroid and contextual features of segmented regions for 11 object classification using a multi-layer perceptron (MLP). In [7], the entropy was used as a texture feature, together with RGB color components and an SVM classifier for detecting vegetation from roadside images. The intensity differences between pixels were combined with a 3D Gaussian model of YUV channels for grass detection [5], and with L, *a*, and *b* components for object segmentation [4]. The motion between video frames estimated by optical flow was also used as a pre-processing step to detect a region of interest [6], from which color and texture features were extracted using a two-dimensional CWT, and also to assist vegetation detection by measuring the resistance of vegetation pixels [13]. In [14], local binary patterns (LBP) and gray-level co-occurrence matrix (GLCM) were combined for discriminating between dense and sparse roadside grasses using a majority voting over three classifiers – support vector machine (SVM), artificial neural network (ANN), and k-nearest neighbour (KNN). In [15], RGB, HLS, and Lab color channels and co-occurrence matrix based texture features were fused for outdoor scene analysis. A set of initial seed pixels was selected based on probabilistic pixel maps built using Gaussian Probability Density Function (PDF) on a selected subset of color and texture features, and pixels were then grown from the initial seeds by integrating region and boundary information in the minimization of a global energy function. In [16], a spatial contextual superpixel model (SCSM) was proposed for roadside vegetation segmentation. Pixel patch selective features were first generated from pixel-level color and patch-level color moments, and further used to train class-specific ANNs. Based on class probabilities produced by ANNs, a superpixel merging strategy was proposed to progressively merge superpixels with low probabilities into the most similar neighbors by performing a double-check on whether a superpixel and its neighour accept each other, as well as enhancing a global contextual constraint. The SCSM produced more than 90% accuracy for a binary classification of seven objects on a cropped roadside dataset.

Most existing visible approaches focus on a binary classification of vegetation vs. non-vegetation. Although different types of color and texture features have been used previously, there is not yet a common feature set that is widely accepted as being capable of working well in natural conditions.

**Table 1** Summary of typical visible approaches for vegetation segmentation.

| Ref. | Color | Texture | Classifier | Class | Data | Result |
|---|---|---|---|---|---|---|
| [12] | RGB, O1, O2, R-G, (R+G)/2-B | Gabor filter, shape | SOM + MLP | Veg, sky, road, wall etc. | 3751 R | 61.1% 80% |
| [7] | RGB | Entropy | RBF SVM + MO | Veg vs. non-Veg | 270 I | 95.0% |
| [6] | RGB, HSV, YUV, CIELab | 2D CWT | SVM + MO | Veg vs. non-Veg | 270 I | 96.1% |
| [5] | YUV (3D Gaussian) | PID | Soft segmentation | Grass vs. non-grass | 62 I | 91% |
| [8] | O1, O2 | NDVI & MNDVI, Gabor filter | Spreading rule | Veg vs. non-Veg | 2000 I 10 V | 95% |
| [17] | H, S | Height of grass (Ladar) | RBF SVM | Grass vs. non-grass | N/A | N/A |
| [4] | Lab | PID | K-means clustering | Object segmentation | N/A | 79% |
| [10] | Gray | Intensity mean & variance, binary edge, neighborhood centroid | Clustering | Grass vs. artificial texture | 40 R | 95% 90% |
| [14] | Gray | LBP, GLCM | SVM, ANN, KNN | Dense vs. sparse grass | 110 I | 92.7% |
| [15] | RGB, HLS, Lab | Co-occurrence matrix | Gaussian PDF + global energy | 5, 5 and 7 objects | 41 I 87 I 100 I | 89.9% 90.0% 86.8% |
| [16] | RGB, Lab | Color moment | Superpixel merging | 7 objects | 650 I 50 I | >90% 77% |

Note: N/A – Not Available, PID - Pixel Intensity Difference, CWT- Continuous Wavelet Transform, LBP - Local Binary Patterns, GLCM - Gray-Level Co-occurrence Matrix, MO - Morphological Opening, SVM - Support Vector Machine, SOM – Self-Organising Map, RBF - Radial Basis Function, NN - Neural Network, KNN - K-Nearest Neighbour, Veg - Vegetation, non-Veg - non Vegetation, I - Image, V - Video, R - Region.

*2.2 Object Segmentation Approach*

Object segmentation approaches aim to find the region where a specific object is present. They often share similar concepts with scene labelling, object categorization, semantic segmentation, etc. in computer vision tasks, and can be potentially applied into roadside vegetation segmentation. According to the techniques and features used, existing approaches can be divided into different categories, such as parametric vs. non-parametric, supervised vs. unsupervised, and pixel based vs. region based.

Early approaches on object segmentation obtain class labels for image pixels using a set of low-level visual features extracted at each pixel [15], [18] or a local patch around each pixel [2]. However, because pixel-level features treat each pixel individually, they are unable to capture statistical characteristics of objects in local regions. While patch-level features are able to capture regional statistic features, they are still prone to noise from background objects due to the difficulty of accurate segmentation of object boundaries. Recent advances [19], [20], [21], [22] have shifted to the adoption of superpixel-level features. Superpixels, which group pixels into perceptually meaningful atomic regions, have advantages of coherent support regions for a single labelling on a naturally adaptive domain rather than on a fixed window, and more consistent statistic feature extraction capturing contextual neighboring information by pooling over feature responses from multiple pixels. The most widely adopted superpixel-level features include color (e.g. RGB [23], [24] and CIELab [18], [19], [23], [25], [26]), texture (e.g. SIFT [24], [27], texton [18], [24], Gaussian filter [23], Gist [25] and PHOG [25]), appearance (e.g. color thumbnail [24]), location [28], shape, etc. For object classification, the vast majority of existing approaches focus on graphical models, such as Conditional Random Fields (CRFs) [22] and Markov Random Field (MRF) [24]. These models often enforce the spatial consistency of category labels between neighboring superpixels (or pixels) by jointly minimizing the total energy of two items - unary potentials which indicate the likelihood of each superpixel (or pixel) belonging to one of semantic categories and pairwise potentials which account for the spatial consistency of category labels between neighboring superpixels (or pixels). However, flat graphical models have limited capacity of capturing higher order context. To address this issue, various local and global contextual features (e.g. relative spatial relationships between objects, such as left, right, top, or bottom locations [31], relative location offsets of objects [32], and object co-occurrence statistics [33]) and hierarchical models (e.g. stacked hierarchical learning [25], pylon model [29], hierarchical reconfigurable template [30], and multi-scale segmentation graph [31]) have been proposed. Hierarchical models generate a pyramid of image superpixels and perform classification optimization over multi-levels of images. One drawback of graphical models is that their parameters are solely learnt from training data, and thus their performance heavily depends on the availability of adequate training data and may not generalize well on testing data. One solution to this problem, particularly for large datasets, is adopting non-parametric approaches [32], which retrieve the most similar training images to a query image and then perform class label transfer from K-nearest neighbours in the retrieval set to the query image. However, non-parametric approaches still depend on the reliability and accuracy of the retrieval strategies.

Recently, deep learning techniques have shown great advantages in extracting discriminative and compact features from raw image pixels rather than using hand-engineered features. The widely used Convolutional Neural Networks (CNNs) utilize convolutional and pooling layers to progressively extract more abstract patterns and demonstrate state-of-the-art performance in many vision tasks [33] including object segmentation. The extracted CNN features can be combined with various classifiers (e.g. MRF, CRFs and SVM) to predict class labels. A representative work is by Farabet et al. [20], which applied hierarchical CNN features into CRFs for class label inference in natural scenes. However, the CRF inference is completely independent from CNN training, and thus Zheng et al. [34] formulated the CRF inference as recurrent neural networks and integrated them in a unified framework. In [35], the recurrent CNN feeds back the outputs of CNNs to the input of another instance of the same network, but it works only on sequential data. Recent extensions to CNN models include AlexNet, VGG-19 net, GoogLeNet, ResNet [36], etc. However, these models may not be directly applicable for our purpose as they often require adequate image resolutions, while the cropped regions in this paper have much low resolutions and substantial variations in the shape and size.

*2.3 Texton based Approach*

Texton [37], which is essentially a set of clustered centres of filter-bank responses or other feature descriptors, has proved as a powerful image representation for generic object segmentation. Its concept is very similar to the bag-of-words (BoW) representation model, and it constructs a vocabulary of visual words for each class or all classes by learning procedures on filter response features from training data, typically with simple but effective K-means clustering, and in the testing, each pixel is then assigned to the nearest-neighbor textons, forming a frequency histogram of those textons which is taken as the image representation. In [38], a universal visual vocabulary of texture textons was built by convolving the image with 17-D filter banks and then aggregating the filter responses over all training images using K-means clustering. Due to high accuracy on texture-rich object categorization, the 17-D filter bank textons were later adopted in many studies [19]. An extension of textons is textonboost [18], which is a boost version of textons to jointly model shape and texture features, where the 17-D filter bank textons are used in conjunction with color features to build a multi-class classifier by iteratively building a strong classifier as a sum of 'weak classifiers' using an adapted version of the Joint Boosting algorithm. Another extension is semantic texton forest [39], which is a bag of semantic textons where a histogram of hierarchical semantic textons is combined with a region prior category distribution to build highly discriminative descriptors for object categorization. A similar texton-based work for outdoor scene analysis to our paper is [4], which constructed textons using K-means clustering from color and texture features comprising of L, *a*, *b* color channels and the L difference between a pixel and its surrounding pixels. A texton-based histogram was then built in a large neighourhood to merge similar clusters based on the Earth Movers Distance for segmenting objects, including vegetation, road, soil and sky etc.

The concept of textons is closely related to dictionary learning, which also received extensive attention in recent years. Xie et al. [40] constructed two dictionary-based representations from CNN models and further combined them with CNN features to form a more powerful hybrid feature representation, achieving state-of-the-art performance for scene recognition. Xie et al. [41] employed an Auto-Encoder network for local descriptor coding and integrated it in a BoW framework for image classification, showing very competitive results. A multiple instance dictionary learning algorithm was presented in [42] for object classification, which projected the data into an embedded feature space and iteratively learnt the dictionary in the spaces of positive and negative bags of data using kernel learning. Zhang et al. [43] provided a review on sparse representation based dictionary learning approaches.

Most existing approaches build generic textons for all classes from the training data and then map features into the closest texton, forming a histogram representation for the image. However, generic textons may not be effective to capture the specific characteristics of each class and handle the confusion between classes with similar characteristics. A histogram representation may fail for small images due to a sparse bin problem. It is worth noting that texton features have seldom been used for vegetation segmentation previously.

This paper investigates the use of superpixel based class-semantic texton occurrences for natural roadside vegetation segmentation. It significantly extends our previous work [44] with comprehensive description of related work, technology, performance evaluation, as well as real-world application. A similar pervious work to ours is [3], which built a set of semantic SIFT words for each of five objects (i.e. sky, tree, building, road and car) and then integrated all words over the image for scene understanding. The visual words were formed by performing K-means clustering on scale invariant feature transform (SIFT) features extracted from local image bocks. The SIFT features in testing images are mapped to the semantic visual words and object classification is accomplished by performing a majority voting on the number of mapped visual words in an image over all classes. But the approach did not consider color textons and assumed each cropped testing image belonging to only one object. The evaluation was limited to only a few manually cropped images. These issues are addressed in our paper.

*2.4 Contributions of This Paper*

The main contributions of this paper are comprised of:

a) We propose novel color-texture class-semantic textons, which take into account not only color and texture features, but also individual semantic characteristics in each class, and thus they retain a high discriminative capacity between classes. The color and texture textons are further integrated to make a collective decision based on the texton occurrences of a pool of pixels within each superpixel, achieving high accuracy for object segmentation.

b) We conduct comprehensive experiments to investigate the impact of key parameters on the performance of the proposed approach. We further demonstrate the feasibility of applying the approach into the practice of vegetation segmentation on real-world video data.

c) We create two natural roadside datasets and make them publicly accessible to other researchers. To the best of our knowledge, they are the first public natural roadside datasets collected from realistic scenarios specifically for vegetation segmentation in this field. They can serve as benchmarking datasets for performance comparisons between relevant algorithms and systems.

**3. Proposed Approach**

This section introduces the systematic framework of the proposed approach that builds class semantic color-texture textons and aggregates texton occurrences over all pixels in each superpixel for a collective decision on vegetation segmentation.

*3.1 Framework of Approach*

Fig. 1 depicts the framework of the proposed approach, comprising of a training stage and a testing stage. During training, an equal set of local regions is manually cropped from the training images for each class. Color and filter bank responses are then extracted from those regions, which are further input into K-means clustering to generate two individual sets of class-semantic color and texture textons for each class. Each set of color or texture textons is combined for all classes to form two class-semantic texton matrices – one for color and one for texture. At the testing stage, the input image is first segmented into a group of heterogeneous superpixels, and color and filter bank features of all pixels in each superpixel are then extracted and mapped separately into one of the learnt color or texture textons using the closest Euclidean distance. The superpixel based color and texure texton occurrences for each class can be obtained and further combined using a linear mixing method. Finally, the segmentation is achieved by assigning all pixels in each superpixel to a class label which has the maximum combined occurrence across all class categories, including brown grass, green grass, soil, road, tree leaf, tree stem, and sky.

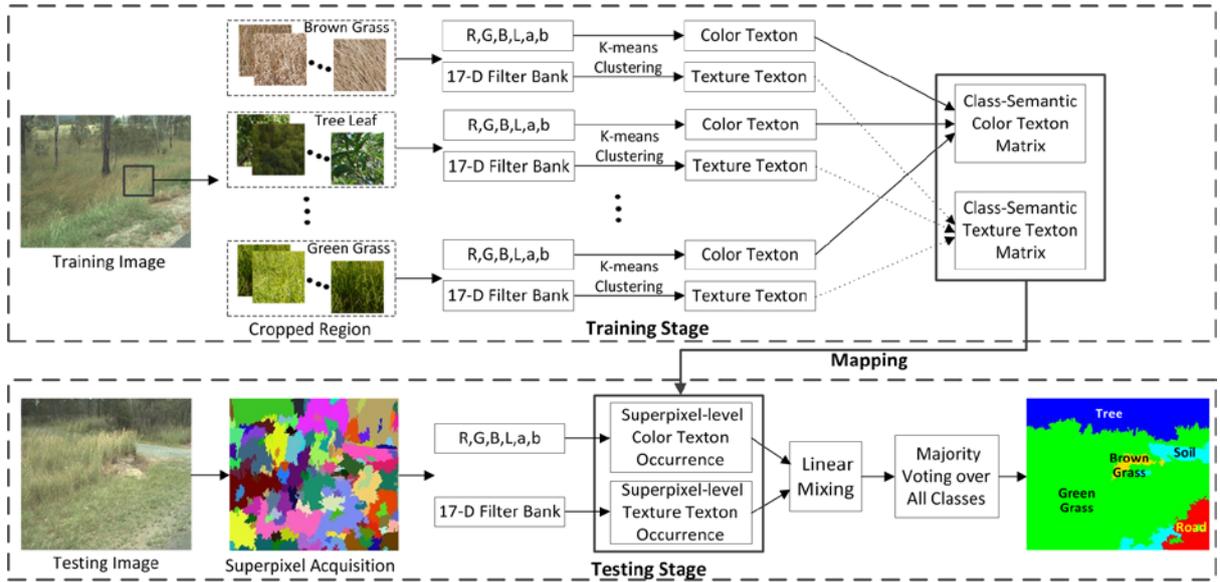

**Fig. 1** Systematic framework of the proposed approach. During training, a set of regions is cropped to generate class-semantic color and texture textons using K-means clustering. During testing, features of all pixels within each superpixel are mapped to one of the learnt textons, and further aggregated into texton occurrences. Superpixel based segmentation is achieved by taking the class with the maximum combined occurrence of color and texture textons.

### 3.2 Superpixel Acquisition

The first task of the proposed approach is to oversegment an input testing image into a set of local superpixels with homogenous appearance, each of which is expected to belong to the same class category. The proposed approach takes superpixels as the basic processing unit due to their compelling characteristics, including simplifying the classification problem from millions of pixels into hundreds of superpixels, utilizing collective decisions over a pool of pixels within each superpixel, significantly reducing the computational complexity, etc. Thus superpixel based classification is expected to be able to significantly reduce the complexity of the categorization process in the proposed approach. There are a lot of popular region segmentation algorithms, such as mean shift [45], JSEG [46] and superpixel [47], and we employ a popular graph-based algorithm [48] due to its high reported performance for natural scene analysis.

*3.3 Color and Texture Feature Extraction*

Feature extraction aims to extract a set of most discriminative features to distinguish the appearance of different object categories. The generation of textons in the proposed approach is based on two types of features: color and texture, which are expected to be able to complement each other for a more effective representation of the most discriminative features of objects.

Color: how to choose a suitable color space is still an open question. One criterion is that the color space should be perceptually consistent with human vision, as the human eye is very adept about distinguishing one object from others even under extremely challenging environmental conditions. This paper adopts the CIELab, which has high perceptually consistency with human vision and demonstrated a good generalized performance on scene understanding [39]. We also include RGB as it may contain complementary information for specific objects. For a pixel at the coordinate (*x*,*y*) in an image, its color feature vector is composed of :

$$V_{x,y}^c = <R, G, B, L, a, b> \qquad (1)$$

Texture: many types of filters have been proposed previously for generating textons for object classification, including Leung and Malik (LM) with 48 filters, Schmid (S) set with 13 filters, maximum response (MR8) set with 38 filters, and Gabor set with a certain number of filters etc. This paper uses 17-D filter banks firstly adopted in [38], which have excellent performance for generic object classification and been widely adopted in existing studies. The 17-D filter banks include Gaussians with 3 different scales (1, 2, 4) applied to L, *a*, and *b* channels, Lapacians of Gaussians with 4 different scales (1, 2, 4, 8), and the derivatives of Gaussians with two different scales (2, 4) for each axis (*x* and *y*) on the L channel. By convolving each image with the filter banks, 17 response images can be obtained and each pixel is characterized by 17 responses. For a pixel at (*x*,*y*) in an image, its texture feature vector is composed of:

$$V_{x,y}^t = <G_{1,2,4}^L, G_{1,2,4}^a, G_{1,2,4}^b, LOG_{1,2,4,8}^L, DOG_{2,4,x}^L DOG_{2,4,y}^L> \qquad (2)$$

*3.4 Class-Semantic Color-Texture Texton Construction*

After obtaining the color and texture features, we proceed to generate two individual sets of middle-level texton features from each of them. Unlike existing texton based approaches that generate a universal visual vocabulary for all classes, this paper extracts a representative set of the most discriminative textons specifically for each class, i.e. class-semantic textons, which are expected to represent more separable and less redundant characteristics for each object to reduce confusion between classes.

Assume there are *C* classes and *n* training pixels in the *i*-th class (*i*=1, 2,…, *C*). Let $V_i^c$ and $V_i^t$ be the color and texture features respectively for the *i*-th class, the K-means clustering algorithm is used to learn a set of textons for each of $V_i^c$ and $V_i^t$ by minimizing:

$$J_c = \sum_{j=1}^{n} \min_k |V_{i,j}^c - T_{i,k}^c|^2 \tag{3}$$

where, $V_{i,j}^c$ is color features of the $j$-th pixel in $V_i^c$, $T_{i,k}^c$ is the $k$-th color textons learnt ($k=1,2,…,K$) for the $i$-th class, and $J_c$ is the error function. The function for texture features is similar to (3) and thus not shown. The $i$-th class-semantic color and texture texton vectors are respectively composed of:

$$T_i^c = <T_{i,1}^c, T_{i,2}^c, …, T_{i,K}^c> \tag{4}$$

$$T_i^t = <T_{i,1}^t, T_{i,2}^t, …, T_{i,K}^t> \tag{5}$$

The textons are basically the cluster centres of color or texture features that have the minimum distance (i.e. Euclidean) between them and all feature descriptors. The value of $K$ controls the number of learnt textons and determines the size of the texton feature space, which often have significant impact on the effectiveness of the learnt textons in representing the characteristics of each class.

Combine the color or texture texton vector for all $C$ classes, a color and a texture texton matrix can be formed respectively:

$$T^c = \begin{Bmatrix} T_{1,1}^c, T_{1,2}^c, …, T_{1,K}^c \\ T_{2,1}^c, T_{2,2}^c, …, T_{2,K}^c \\ \vdots \\ T_{C,1}^c, T_{C,2}^c, …, T_{C,K}^c \end{Bmatrix} \text{ and } T^t = \begin{Bmatrix} T_{1,1}^t, T_{1,2}^t, …, T_{1,K}^t \\ T_{2,1}^t, T_{2,2}^t, …, T_{2,K}^t \\ \vdots \\ T_{C,1}^t, T_{C,2}^t, …, T_{C,K}^t \end{Bmatrix} \tag{6}$$

The above two matrices are comprised of color and texture textons for all $C$ classes learnt from training data respectively, and are expected to contain representative and discriminative features for each class, which are used for distinguishing between objects in testing data.

*3.5 Superpixel based Texton Occurrence and Vegetation Segmentation*

For all pixels $P^I$ in an image and given a set of object categories $C^N$, the task of object segmentation is to find a mapping $M: P^I \rightarrow C^N$ so that each pixel corresponds to a category. Given the learnt color and texture texton matrices, this part proposes a majority voting classification strategy to obtain the class label for all pixels in a testing image using superpixel based texton occurrences, which essentially aggregates texton occurrences over all pixels in each superpixel to derive a collective classification decision. To be specific, we first map all pixels in a testing image into one of the learnt color textons and one of the learnt texture textons respectively, and then calculate the occurrences of mapped color and texture textons of all pixels in each superpixel for each class. The color and texture texton occurrences are further combined using a linear mixing method to obtain the class probabilities for each superpixel which indicate the likelihood of this superpixel belonging to each of all classes, and all pixels in this superpixel are finally assigned to the class label with the highest probabilities across all classes.

For an input image $I$, it is first segmented into a set of superpixels with homogenous features using a popular graph-based algorithm [48]:

$$S = <S_1, S_2, …, S_L> \tag{7}$$

where, $L$ is the number of segmented superpixels, and $S_l$ stands for the $l$-th superpixel.

Assume there are $m$ pixels in the superpixel $S_l$, the color and texture feature vectors of $S_l$ can be extracted using Equations (1) and (2) respectively, i.e. $V^c_{S_l} = \bigcup_{x,y \in S_l} V^c_{x,y}$ and $V^t_{S_l} = \bigcup_{x,y \in S_l} V^t_{x,y}$. The $V^c_{S_l}$ and $V^t_{S_l}$ are mapped to the learnt class-semantic color and texture textons respectively by finding the closest texton using a distance metric (i.e. Euclidean):

$$f(V^c_{x,y}, T^c_{i,k}) = \begin{cases} 1, & if \ \|V^c_{x,y} - T^c_{i,k}\| = \min_{q=1,2,...C; p=1,2,...,K} \|V^c_{x,y} - T^c_{q,p}\| \\ 0, & otherwise \end{cases} \quad (8)$$

A color texton occurrence matrix which accounts for the number of the mapped textons $T^c_{i,k}$ for all pixels in $S_l$ can be obtained using:

$$A^c_{i,k}(S_l) = \sum_{x,y \in S_l} f(V^c_{x,y}, T^c_{i,k}) \quad (9)$$

The values in the color texton occurrence matrix are accumulated for the $i$-th class, yielding the occurrence of color textons in the superipixel $S_l$ for this class:

$$A^c_i(S_l) = \sum_{k=1}^{K} A^c_{i,k}(S_l) \quad (10)$$

Repeat the above procedure for the texture feature vector $V^t_{S_l}$ to obtain the occurrence of texture textons in $S_l$ for the $i$-th class:

$$A^t_i(S_l) = \sum_{k=1}^{K} A^t_{i,k}(S_l) \quad (11)$$

Combine the occurrences of color and texture textons using a simple linear mixing method to generate a combined occurrence in $S_l$ for the $i$-th class:

$$A^l_i = A^c_i(S_l) + w * A^t_i(S_l) \quad (12)$$

where, $w$ is a weight for texture textons relative to a fixed value of 1 for color textons, and it indicates the relative contribution of texture textons to the combined results. The combined occurrence is further converted into a class probability by dividing the total number (i.e. $m$) of all pixels within $S_l$:

$$p^l_i = A^l_i / m \quad (13)$$

A class probability vector of $S_l$ for all classes can be obtained:

$$P^l = <p^l_1, p^l_2, ..., p^l_C> \quad (14)$$

All pixels in $S_l$ are finally assigned to the $c$-th class which has the maximum probability over all classes:

$$S_l \in c-\text{th class if } p^l_c = \max_{i=1,2,...C} p^l_i \quad (15)$$

The above procedure makes a collective classification decision for each superpixel based on color and texture texton occurrences of all pixels within the superpixel, to utilize supportive information in a spatial neighbourhood. Thus, the result is expected to be robust to small error or noise in the superpixel. Note that the pre-processing step of image segmentation is not performed for manually cropped regions, which have only one object in each, and each of them is treated as an individual superpixel.

## 4. Experiments

We evaluate the performance of the proposed approach on a cropped region dataset and a natural roadside image dataset. We also comparatively evaluate varied values of several key parameters to achieve an optimal approach with balanced accuracy and computation, and further apply the approach into the practice of vegetation detection on real-world video data.

*4.1 Datasets*

The datasets used for the experiment come from the Department of Transport and Main Roads (DTMR), Queensland, Australia. The DTMR has been collecting roadside video data each year using cameras mounted on a vehicle driving across state roads in Queensland. There are four cameras mounted in the front, left, right and rear part of the vehicle respectively, to capture video data from four directions. All video data is in an AVI format with a frame resolution of 1632×1248 pixels. The data used in this paper is selected from the video captured using the left camera, which focuses on vegetation regions. From the DTMR data, two image datasets were created, which are publicly accessible at https://sites.google.com/site/cqucins/projects. Note that no specific validation set was separately created for the two datasets.

a) Cropped region dataset. We manually crop a total of 650 small regions from 230 original frames (samples shown in Fig. 2) for seven types of objects (100 images per object except 50 for sky, due to relatively less appearance variations in sky regions), including brown grass, green grass, tree leaf, tree stem, soil, road and sky as shown in Fig. 3. Each cropped region contains only one type of object, and has different texture or structure from other regions. The cropped dataset allows manual assistance to cover as many as possible types of appearance variations of each object in real-world situations, which helps to build class-semantic textons.

b) Natural image dataset. We manually select 50 images from video data (independent from those used for the cropped dataset). These images were selected to be representative of different real-life cases, covering various types of vegetation and other objects, such as soil, road, and sky. All pixels were manually annotated into seven categories of objects, including brown grass, green grass, tree, soil, road, sky, and unknown objects, and they serve as ground truths for performance evaluations. Note that tree leaf and stem are combined into one category of tree.

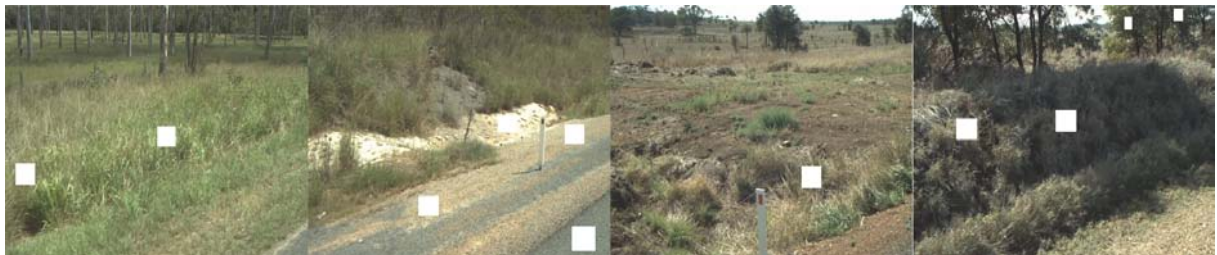

**Fig. 2** Samples of DTMR video frames used for creating the cropped region data. The while rectangles in the figures provide an indication of the location where the cropped regions are extracted and their sizes and shapes. Note that only a proportion of representative cropped regions are included in the final dataset.

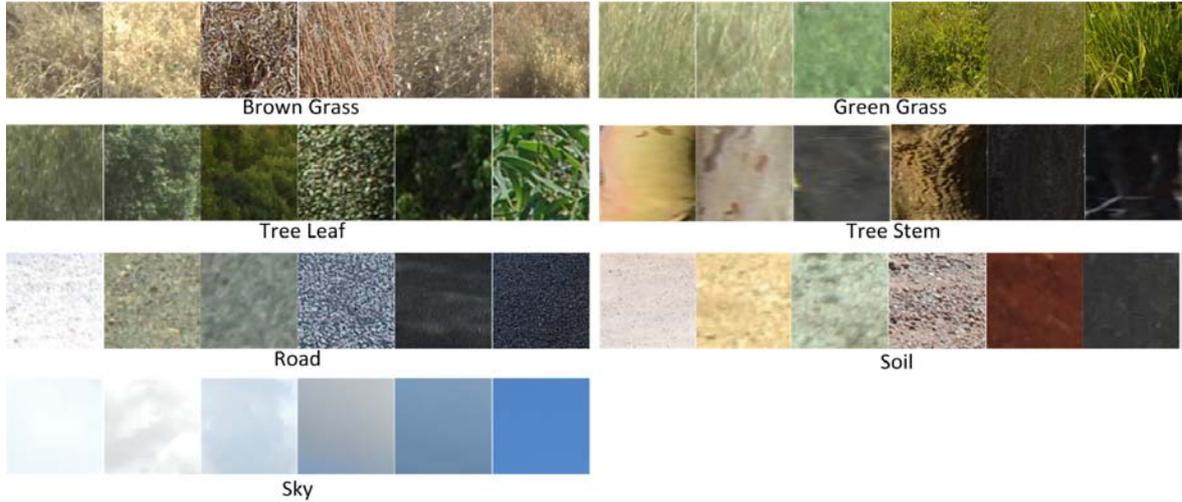

**Fig. 3** Samples of cropped regions for seven types of objects. Note the big variations in the appearance of the same object, and the high similarity between some objects (e.g. green grass and tree leaf, soil and road), which pose great challenges for accurate classification. These regions have varying resolutions and shapes, and blurred content.

*4.2  Implementation Details and Parameter Setting*

All natural images are scaled to a fixed size of 320×240 pixels to facilitate the processing of image segmentation and reduce the computational cost. The parameters of the graph-based image segmentation algorithm are set based on the recommended settings in [49], i.e. $\sigma = 0.5$, $k = 80$, and $min = 80$ for an image size of 320×240 pixels. To ensure balanced training data for each object, 120 pixels are selected from random coordinates in each cropped region and used for generating the color-texture textons using the K-means clustering algorithm. The number of color and texture textons is set to be the same during the fusion of color and texture textons, i.e. color-texture textons, on both cropped region and natural image datasets. The whole system was implemented under a Matlab platform using a Macbook laptop with a configuration of 1.8 GHz Intel Core i5 processor and 4GB memory.

**Evaluation metrics.** The performance of the proposed approach is evaluated using two measurements: *global accuracy* measured in terms of all pixels across all testing images and all classes and *class accuracy* averaged over each class using pixelwise comparisons between classified results and ground truths. The global accuracy is biased favourably to frequently occurring classes, and pays less attention to the classes with low frequency. In contrast, the class accuracy treats the classification of each class equally regardless of their occurring frequency. Thus, they are able to reflect different aspects of the performance. Four cross-validations are used to obtain an average accuracy. In details, the cropped regions for each class are split into four equivalent subsets, and for each cross-validation, three subsets are used for training and the left one for testing.

*4.3  Global Accuracy vs. Number of Textons*

Fig. 4 shows the global accuracy of the proposed approach versus the number of textons in classifying seven object classes on the cropped region dataset, and classifying six objects on the natural image dataset. Three types of features are compared, including color-texture textons, color textons alone, and texture textons alone, as well as two types of classification strategies: superpixel based collective decision (i.e. superpixel) and pixel based single decision (i.e. pixel). Note that color-texture textons indicate the combination of an equal number of color and texture textons.

We can observe that superpixel based classification has much higher accuracy (about 14%) than pixel based classification for both the cases of using color or texture textons on the two datasets. This proves the benefit of aggregating collective classification decisions over a pool of pixels within each superpixel, which leads to more robust results than pixel based classification. For both superpixel and pixel based classification and both datasets, color textons and texture textons exhibit a similar overall performance. Their performance increase gradually along with a larger number of textons on the region dataset, but tend to level off on the image dataset. For both datasets, fusion of color and texture textons (i.e. color-texture textons) leads to slightly higher global accuracy than using color or texture textons alone. The highest global accuracy of 79.9% is obtained for the region dataset using 90 color-texture textons, and the highest accuracy of 74.2% for the image dataset using 30 color-texture textons. Another advantage of using color-texture textons, compared to using color or texture textons, is that higher global accuracy can be obtained using a small number of textons, which may be important for applications where a real-time computational speed is critical.

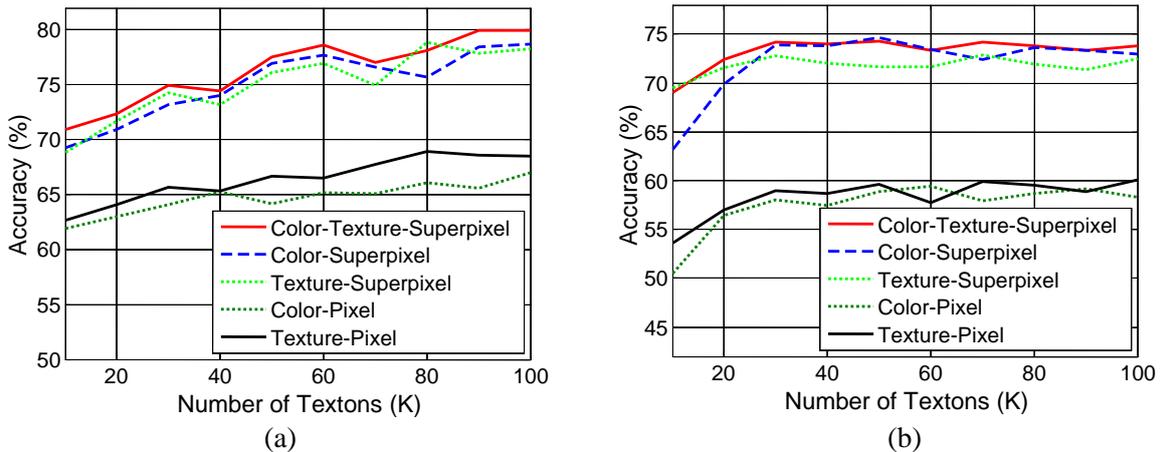

**Fig. 4** Global accuracy vs. the number of textons on the (a) cropped region and (b) natural image datasets. Five approaches using color-texure textons, color textons, and texture textons with pixel or superpixel based classification are compared. Parameters: the number (i.e. $K$) of color and texture textons is the same, the combination weight for texture textons $w$=1, the size of Gaussian filters: 7×7 pixels, and distance metric: Euclidean.

Fig. 5 reveals the computation performance versus the number of color-texture textons on the two datasets. The overall computation is the average seconds required for each testing image (or region), which is primarily comprised of two processing stages: feature extraction and texton mapping and classification (i.e. texton M&C). For both datasets, there exhibits an approximately linear relationship between the overall computation and the number of textons, and the vast majority of the overall time is used for texton mapping and classification. By contrast, the time used for color and texture feature extraction remains constant and takes only a small proportion of the overall time. Note that natural images require more time than cropped regions due to their higher resolutions. Thus, there is a necessity of choosing a suitable number of color-texture textons to achieve a good balance between accuracy and computation. This paper chooses to use 60 and 30 color-texture textons for the region and the image datasets respectively, for which the accuracy is 78.6%, and 74.2%, and the computational time is 1.3 and 2.6 seconds respectively.

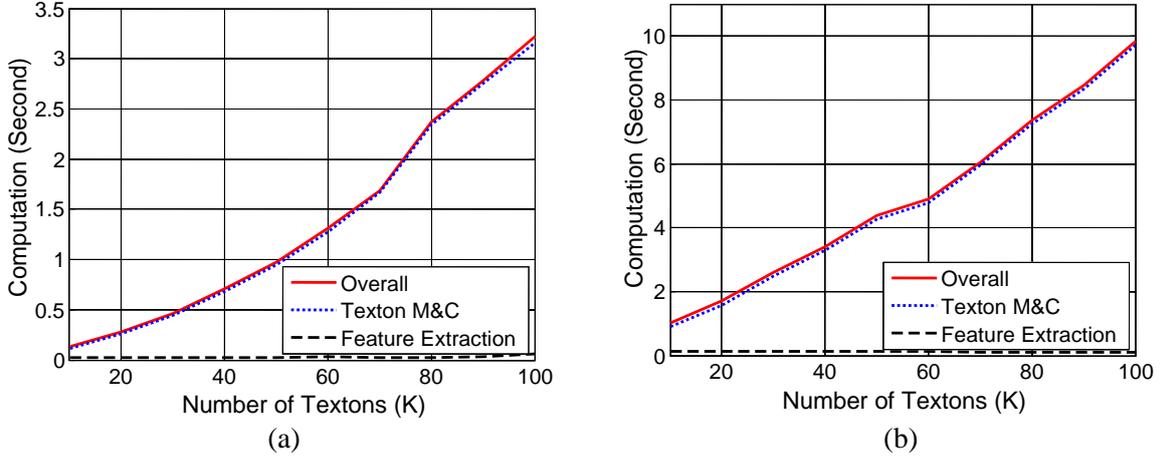

**Fig. 5** Computational performance vs. the number of textons on the (a) cropped region and (b) natural image datasets. The time is the average seconds per image (or region) required for calculating color and texture features (i.e. Feature Extraction), and performing texton mapping and classification (i.e. Texton M&C). The overall time is the total seconds required for the classifying each image (or region) using the proposed approach.

*4.4 Global Accuracy vs. Combination Weights*

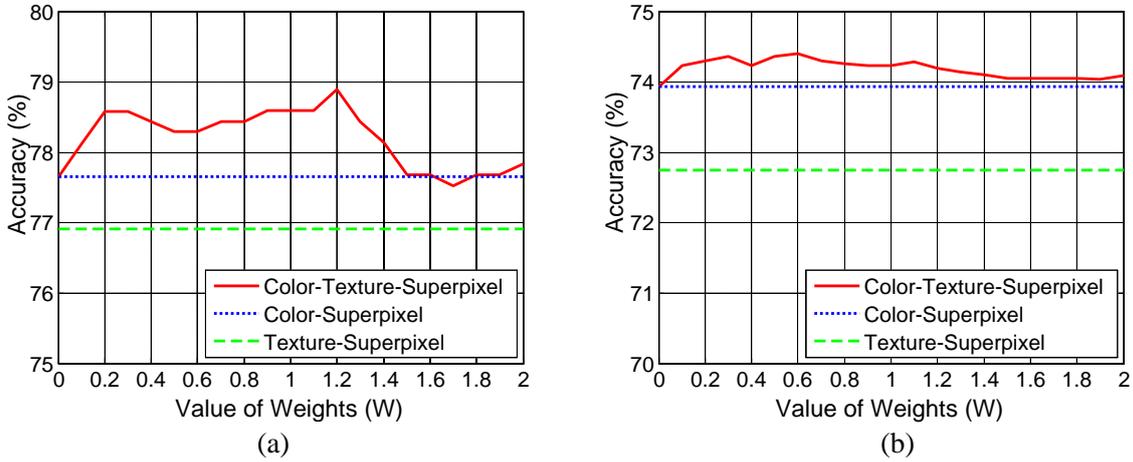

**Fig. 6** Global accuracy vs. the value of combination weights on the (a) cropped region and (b) natural image datasets. The weight of texture textons is a value relative to a fixed value of 1 for color textons. Color & texture texton no. = 60 (region dataset) and 30 (image dataset); the size of Gaussian filters: 7×7 pixels; distance metric: Euclidean.

We examine the impact of the weight given to texture textons in their combination with color textons on the global accuracy. The weight (i.e. $w$ in (12)) is a value relative to a fixed value of 1 for color textons, and it indicates the contribution of texture textons to the combined results. Fig. 6 shows the results of the proposed approach in terms of global accuracy using a *w* value ranging from [0.1, 1.5], and the results are obtained using 60 and 30 color-texture textons on the cropped region and natural image datasets, respectively. It can be seen that, for both the datasets, fusion of color and texture textons has higher global accuracy than using color or texture textons alone for most values of the weight, and the best overall performance is obtained using weight values between 0.2 and 1.2. This may indicate that color textons play a slightly more significant role than texture textons in the proposed approach. The highest accuracy of 78.9% is achieved for the cropped region dataset when *w* is 1.2 and 74.4% for the natural image dataset when *w* is 0.6.

### 4.5 Global Accuracy vs. Size of Gaussian filters

We also investigate the impact of the size of Gaussian filters on the global accuracy as shown in Fig. 7. The size values determine the range of the spatial neighborhood from which texture features are extracted using Gaussian filters, and thus they may have substantial impact on the effectiveness of the generated texture textons in representing discriminative characteristics for each object. The performance of three approaches – superpixel based classification using color-texture textons, pixel or superpixel based classification using texture textons are compared using five sizes of Gaussian filters, ranging from 5×5 to 15×15 pixels with an interval of 2 pixels in the width and the height. The results are obtained using 60 and 30 color-texture textons on the region and the image datasets, respectively. For all the three methods and both datasets, there are only small differences in the accuracy using five sizes of Gaussian filters, but small sizes appear to slightly outperform higher sizes, particularly for the region dataset. The highest accuracies of 78.9% and 74.6% are achieved using the sizes of 7×7 and 9×9 respectively for the two datasets. Considering the fact there is little difference between the performance of using the sizes 7×7 and 9×9 on the image dataset, and thus a size of 7×7 is used for both datasets in this paper.

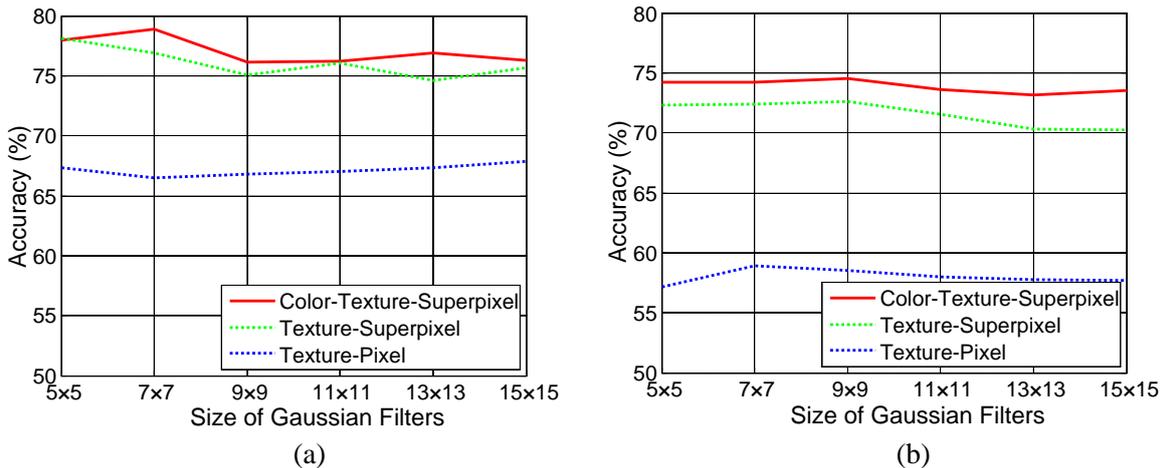

(a)  (b)

**Fig. 7** Global accuracy vs. the size of Gaussian filters on the (a) cropped region and (b) natural image datasets. Parameters: the combination weight *w*=1.2 (region dataset), and 1 (image dataset); Color & texture texton no. = 60 (region dataset) and 30 (image dataset); distance metric: Euclidean (for both datasets). Note that the performance of color textons is not impacted by varied sizes of Gaussian filters.

*4.6 Global Accuracy vs. Distance Metrics*

Another important factor impacting the creation of textons is the distance metric used for K-means clustering. The metric determines the measurement unit that is used to calculate the dissimilarity (or distance) between color or texture features of pixels, and thus it has direct influence on the results of color and texture texton construction, and texton mapping. Table 2 compares the results of four distance metrics - Squared Euclidean, Sum of absolute differences (CityBlock), Cosine, and Correlation. From the table, we can observe that the use of different distances exerts only small influence on the performance of using color-texture textons or color textons, but larger influence on that of using texture textons. Among the four metrics, the Euclidean has the highest accuracy for both color-texture textons and texture textons, whereas the Cityblock produces the highest accuracy for texture textons on the two datasets. We further investigate whether there is any improvement to the performance of color-texture textons when the Euclidean is used for generating color textons and the Cityblock for generating texture textons. Our results show 77.5% accuracy on the region dataset, which is lower than the results of using the Euclidean for both color and texture textons.

**Table 2** Global accuracy (% ± standard deviation) of the proposed approach vs. the type of distances used for generating color and texture textons, as well as texton mapping. The size of Gaussian filters: 7×7 pixels. For the region dataset, the combination weight *w*=1.2 and color & texture texton no. = 60. For the image dataset, *w*=1 and color & texture texton no. = 30.

|  | **Texton** | **Euclidean** | **CityBlock** | **Cosine** | **Correlation** |
|---|---|---|---|---|---|
| Cropped Region | Color-Texture | **78.9**±3.4 | 75.5±2.2 | 78.0±5.0 | 78.1±2.2 |
|  | Color | **77.7**±2.3 | 73.5±3.3 | 76.8±3.5 | 77.5±3.2 |
|  | Texture | 76.9±4.4 | **78.1**±3.8 | 71.9±5.3 | 72.5±3.7 |
| Natural Image | Color-Texture | **74.2** | 74.2 | 72.1 | 72.3 |
|  | Color | **73.9** | 72.9 | 71.8 | 72.0 |
|  | Texture | 72.8 | **73.3** | 65.1 | 66.6 |

*4.7 Class Accuracy and Confusion Analysis*

Table 3 compares the class accuracy of the proposed approach to five other approaches. It can be seen that color-texture textons have the highest global accuracy on both datasets, and highest average class accuracy on the cropped region dataset. Supervised color textons perform better for classifying road pixels than texture textons, while texture textons are better for classifying soil and sky pixels on the cropped region dataset. This result, however, is not observed on the image dataset where color textons show higher class accuracies than texture textons for all objects, except for brown grass. Compared to superpixel based classification, there are significant reductions on the class accuracy for all objects on both datasets using pixel based classification, particularly for brown grass and road which have nearly 20% reductions. This indicates the importance of utilizing a superpixel based collective decision rather than a pixel based single decision for classifying objects.

**Table 3** Comparisons of class accuracy (%) of the proposed approach to other approaches.

(a) Cropped region dataset (color-texture texton no. = 60, 7×7 Gaussian filters, $w$=1.2, Euclidean distance)

|  |  | BG | GG | Road | Soil | TL | TS | Sky | Average | Global |
|---|---|---|---|---|---|---|---|---|---|---|
| Superpixel based | Color-Texture | 86.2 | 88.0 | 84.2 | 57.0 | 79.0 | 68.0 | 96.1 | **79.8** | **78.9** |
|  | Color | 85.2 | 87.0 | 90.1 | 51.0 | 80.0 | 65.0 | 92.2 | 78.6 | 77.6 |
|  | Texture | 85.2 | 88.0 | 80.2 | 59.0 | 76.0 | 62.0 | 98.0 | 78.3 | 76.9 |
| Pixel based | Color | 56.0 | 65.3 | 66.5 | 48.7 | 58.5 | 56.6 | 87.8 | 62.8 | 65.0 |
|  | Texture | 52.4 | 71.3 | 68.0 | 44.4 | 55.3 | 55.1 | 91.4 | 62.6 | 67.2 |

(b) Natural image dataset (color-texture texton no. = 30, 7×7 Gaussian filters, $w$=1, Euclidean distance)

|  |  | BG | GG | Road | Soil | Tree | Sky | Average | Global |
|---|---|---|---|---|---|---|---|---|---|
| Superpixel based | Color-Texture | 73.5 | 78.7 | 85.7 | 42.5 | 67.6 | 96.6 | 74.1 | **74.2** |
|  | Color | 71.3 | 81.7 | 83.9 | 44.4 | 67.9 | 98.1 | **74.6** | 73.9 |
|  | Texture | 74.2 | 73.7 | 83.7 | 36.3 | 65.5 | 94.1 | 71.3 | 72.8 |
| Pixel based | Color | 47.4 | 65.1 | 66.9 | 41.0 | 61.7 | 94.0 | 62.7 | 58.0 |
|  | Texture | 54.9 | 56.4 | 70.1 | 35.6 | 59.4 | 88.4 | 60.8 | 59.0 |

Note: BG – brown grass; GG – green grass; TL – tree leaf; TS – tree stem.

Table 4 displays the confusion matrix of seven or six objects using color-texture textons. For both datasets, sky is the easiest one for correct classification with more than 96% accuracies, and road is also classified with high accuracies. The results agree with those in [15] and [2], where sky and road also have the two highest classification accuracies among five objects on the OU and MA datasets, and among eight objects on a road scene video dataset, respectively. By contrast, soil is the most difficult one with only 57.0% and 42.5% accuracies respectively on the two datasets, and a significant proportion (more than 33%) of soil pixels are misclassified as brown grass, probably due to the similarity of a yellow color between them. In addition, more than 17% tree pixels are misclassified as road. A similar result was also observed in [15] that the top of some tree leaves were wrongly recognised as road. The results imply the necessity of adopting more discriminative texture features that are specifically designed to distinguish between them under natural conditions. Unlike the result on the region dataset that there is little confusion between brown and green grasses, brown and green grass pixels are also prone to be misclassified to each other on the image dataset, indicating typical challenges for robust vegetation segmentation in natural images that are not observed in manually cropped data.

**Table 4** Confusion matrix for different classes using the proposed approach.

(a) Cropped region dataset

|  | BG | GG | Road | Soil | TL | TS | Sky |
|---|---|---|---|---|---|---|---|
| BG | **86.2** | 0 | 1.0 | 6.9 | 0 | 5.9 | 0 |
| GG | 1.0 | **88.0** | 0 | 0 | 11.0 | 0 | 0 |
| Road | 0 | 0 | **84.2** | 1.0 | 0 | 11.8 | 3.0 |
| Soil | 33.0 | 0 | 3.0 | **57.0** | 0 | 4.0 | 3.0 |
| TL | 1.0 | 11.0 | 1.0 | 0 | **79.0** | 8.0 | 0 |
| TS | 8.0 | 0 | 17.0 | 4.0 | 3.0 | **68.0** | 0 |
| Sky | 0 | 0 | 1.9 | 2.0 | 0 | 0 | **96.1** |

(b) Natural image dataset

|  | BG | GG | Road | Soil | Tree | Sky |
|---|---|---|---|---|---|---|
| BG | **73.5** | 14.7 | 3.3 | 3.1 | 5.4 | 0.0 |
| GG | 7.8 | **78.7** | 2.4 | 0.5 | 8.6 | 0.0 |
| Road | 7.3 | 0.4 | **85.7** | 6.1 | 0.1 | 0.4 |
| Soil | 39.6 | 5.8 | 7.0 | **42.5** | 5.1 | 0.0 |
| Tree | 5.8 | 5.4 | 18.0 | 0.3 | **67.6** | 2.9 |
| Sky | 0.2 | 0.0 | 2.8 | 0.1 | 0.3 | **96.6** |

Fig. 8 displays samples of segmentation results, showing promising overall accuracy. The results on these samples visually confirm the confusion results between objects in Table 4. The confusion between brown and green grasses is partially due to the difficulty of creating clearly distinguishable ground truths for them. Tree pixels are prone to be misclassified as road, due to their similarity in texture and overlap in a dark green color. In addition, soil and brown grass also tend to be misclassified to each other, due to a similar yellow color.

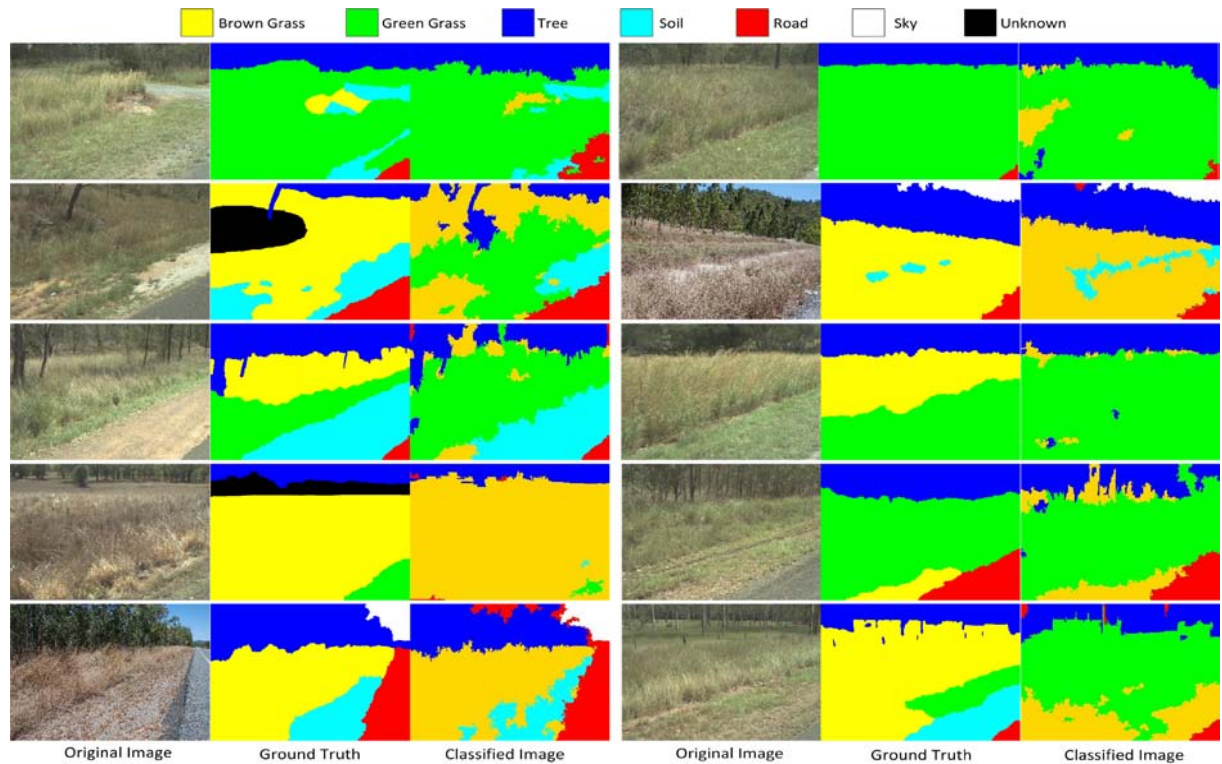

**Fig. 8** Samples of segmentation results on the natural image dataset. The classified results in the right figure are compared with pixelwise ground truths in the middle figure. Note that the yellow color for brown grasses in ground truth images are brighter than that in classified result images.

*4.8 Application to Natural Roadside Video*

We also apply the proposed approach into the practice of vegetation segmentation on a set of 36 videos taken by the DTMR on two state roads within the Fitzroy region, Queensland, Australia[1]. Fig. 9 shows a subset of original frames from three sample videos and their corresponding segmentation results. These frames are manually selected to be representative of different scene content and varying environmental conditions so that they can reveal the performance of the proposed approach under realistic scenarios. It can be seen from the figure that the majority of grass and tree regions are classified successfully, proving the effectiveness and direct applicability of the proposed approach in supporting real-world applications. The results also illustrate a small portion of misclassified pixels, which reveal typical challenges for vegetation segmentation on real-world video data. To be specific, regions accompanied by the shadow of objects which are prone to be misclassified as tree stems, primarily due to their similar characteristic of a dark color. For the same reason, a small proportion of road regions are also misclassified as tree stems. A similar effect of shadows was also observed in [15], where the massive presence of shadows leads to misclassification of tree to unknown objects in outdoor scenes. This indicates the importance of handling the impact of lighting variations for more accurate segmentation on natural data. In addition, there also exist some confusion between soil and brown grass pixels, due to their similarity in a yellow color particularly under a shining condition. The results reflect the fact that color is still the dominant impacting factor in leading to the confusion between objects, and thus it is still necessary to incorporate more effective texture features to further improve the results.

---

[1] The classification results on all videos and detailed information about the collection of these videos will be available at: https://sites.google.com/site/cqucins/projects.

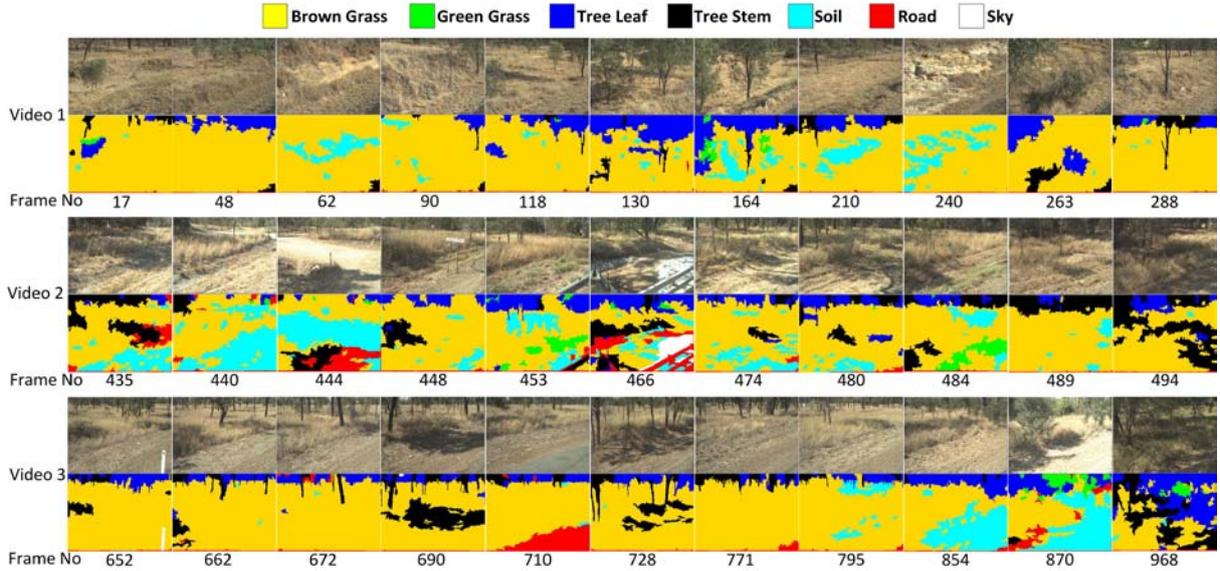

**Fig. 9** Segmentation results on frame samples from three videos. Each row corresponds to one video, and each video contains 1,203 frames in total. Frames in the three rows are selected from the beginning, middle, and end sections of the video, respectively. The frame number in each video is shown below each figure.

*4.9 Performance Comparisons*

To provide a performance comparison with baseline approaches, we first report the performance of the popular CNNs on the cropped region dataset as shown in Table 5. Due to low image resolutions of many regions in the dataset, the LeNet-5 [50], which was initially designed for handwritten recognition, is used here. Note that recently developed CNN models such as AlexNet, VGG-19 and GoogLeNet cannot be directly used here as they were built for larger image resolutions, e.g. 224×224 pixels. To keep a consistent size of input data, all cropped images are resized into $W \times W$ pixels $W \in \{32, 64, 128\}$ using 1) a varied ratio method which resizes both the width and height of regions into $W$ pixels and does not keep the aspect ratio of the width to the height, and 2) a fixed ratio method which resizes the larger one between the width and height of regions into $W$ pixels and keeps the aspect ratio. Studies [33] indicated that keeping the aspect ratio of an image helps preserve the shape of objects and boost the performance. From Table 5, we can see that, for regions resized using varied ratios, the global accuracy drops from 73.5% to 55.6% when the image size increases from 32×32 to 128×128 pixels, and this is probably because using a small size helps prevent substantial information loss due to resizing small regions into a larger size. When fixed ratios are used, using 64×64 pixels produces the highest accuracy of 75.9%, and this is expected as the image resizing is based on the larger value between the width and height. The results also confirm the benefit of keeping aspect ratios of images in minimizing object distortion and achieving higher accuracy. By contrast, using 128×128 pixels has the lowest accuracy for both varied and fixed ratio methods, due to low resolutions of cropped regions.

**Table 5** Global accuracy (%) obtained using the LeNet-5 approach on the cropped region data. The cropped regions are resized using a varied and a fixed ratio of the width to the height, respectively. The evaluations are based on four random cross-validiatons.

| Image size (pixels) | 32×32 | 64×64 | 128×128 |
|---|---|---|---|
| Varied ratio of width to height | 73.5 | 67.9 | 55.6 |
| Fixed ratio of width to height | 64.8 | 75.9 | 34.0 |

Table 6 compares the proposed approach with other approaches on the cropped region dataset and the public Croatia roadside grass dataset [51]. For the cropped region dataset, the proposed approach is compared with four types of approaches: 1) the generic texton histogram approach which creates a universal set of textons for all classes using K-means clustering, and classifies each cropped region to a class which has the nearest distance of texton histograms with this region. For a fair comparison, the same types of color and texture features in this paper are used. 2) The LeNet-5 approach that uses 64×64 regions resized using a fixed ratio method (see Table 5). 3) Pixel characteristic approach [52], which constructs a set of pixel-level statistic color features to represent vegetation characteristics and performs vegetation segmentation using an SVM classifier. 4) Approaches that use three opponent color channels $O_1 O_2 O_3$, their first three moments, and three classifiers – ANN, Linear SVM, and KNN. It can be seen that the proposed approach outperforms all benchmarking approaches and achieves the highest accuracy. The proposed class-semantic textons have significantly higher accuracy (i.e. 19.2%) than using generic textons, confirming the benefit of generating a set of textons specific to each class. Surprisingly, the LeNet-5 produces slightly lower accuracy than the proposed approach, and this may be partly due to information lost in the image resizing process. The proposed approach also achieves 2.9% higher accuracy than the pixel characteristic approach and nearly the same or higher accuracies than the approaches using $O_1 O_2 O_3$ color and moments with an SVM, ANN or KNN classifier. The results indicate a superior performance of our approach compared with state-of-the-art approaches.

The Croatia dataset includes 270 images collected using a right-view camera along public roads, focusing primarily on green grasses and roads. All images have a frame resolution of 1920×1080 pixels and all pixels are manually annotated into two categories of grass and non-grass. Following [6], [7], [51], ten-fold random cross-validations are conducted to obtain an average accuracy. The proposed approach outperforms the three approaches using fusion of five color channels (i.e. BlueSUAB) and 2D CWT texture features, using threshold-based visible vegetation index, and using threshold-based green-red vegetation index respectively in [51], and the approaches using RGB and HSV color features in [7]. Our approach has slightly lower accuracy (i.e. 1.1% and 2.3%) than the approaches using fusion of RGB and entropy [7], and fusion of Lab and 2D CWT features in regions of interest detected using optical flow [6], but its performance is obtained using images with a substantially lower pixel resolution ($320 \times 240$ vs. $1920 \times 1080$ pixels). Thus, the proposed approach is able to produce performance comparable to state-of-the-art approaches using low-resolution data, which is critical for real-time processing in real-world applications.

**Table 6** Performance comparisons with state-of-the-art approaches (color-texture texton no. = 60, 7×7 Gaussian filters, combination weight $w = 1.2$, Euclidean distance).

| Dataset | Approach | Classifier | Object No. | Resolution | Acc. (%) |
| --- | --- | --- | --- | --- | --- |
| Cropped region | Color-texture texton (proposed) | KNN | 7 | - | **78.9** |
| | Generic texton histogram | KNN | 7 | - | 59.7 |
| | LeNet-5 [50] | - | 7 | 64×64 | 75.9 |
| | Color statistic [52] | SVM | 7 | - | 77.0 |
| | $O_1O_2O_3$+color moment [53] | ANN | 6 | - | 79.0 |
| | $O_1O_2O_3$+color moment [53] | SVM | 6 | - | 75.5 |
| | $O_1O_2O_3$+color moment [53] | KNN | 6 | - | 68.6 |
| | $O_1O_2O_3$ [53] | ANN | 6 | - | 72.6 |
| Croatia | Color-texture texton (proposed) | KNN | 2 | 320×240 | **93.8** |
| | BlueSUAB+2D CWT [51] | SVM | 2 | 1920×1080 | 93.3 |
| | Visible vegetation index [51] | Thresholding | 2 | 1920×1080 | 58.3 |
| | Green-red vegetation index [51] | Thresholding | 2 | 1920×1080 | 67.6 |
| | Lab+2D CWT+optical flow [6] | SVM | 2 | 1920×1080 | **96.1** |
| | RGB+entropy [7] | SVM | 2 | 1920×1080 | 94.9 |
| | RGB [7] | SVM | 2 | 1920×1080 | 92.7 |
| | HSV [7] | SVM | 2 | 1920×1080 | 87.3 |

## 5. Discussion

The main lessons learnt from the experimental results in this paper are as follows.

a) Superpixel based classification, which aggregates collective classification decisions over a pool of pixels within each superpixel to utilize supportive information in a spatial neighbourhood, has consistently much higher accuracy than pixel based classification on both the cropped and image datasets. The result has a more general indication that it is beneficial for traditional pixel based object segmentation methods to consider a decision-level fusion at a higher spatial neighbourhood for higher accuracy.

b) A larger number of color and texture textons leads to higher accuracy on the cropped dataset, but the performance tends to level off on the image dataset. For both datasets, color-texture textons achieve slightly higher accuracy than using color or texture textons alone, and they exhibit an approximately linear relationship between the overall computation and the number of textons. Thus, it is important to select a proper number of textons for balanced performance between accuracy and computation.

c) The size of Gaussian filters used for texture feature extraction has small impact on the classification accuracy of the generated texture textons, but small sizes (e.g. 5×5 and 7×7 pixels) appear to slightly outperform higher sizes (e.g. 11×11 and 15×15 pixels) on both datasets. This implies that the most discriminative texture features of objects in the datasets exist in a small spatial neighbourhood, and a larger size may introduce noise into the generated texture textons.

d) A proper value of the weight given to texture textons in their combination with color textons helps to achieve higher accuracy. The best performance is obtained using a weight value between 0.2 and 1.2 for a fixed of 1 for color textons, which indicates color textons playing a slightly more significant role than texture textons in the proposed approach.

e) The distance has small influence on the performance of using color-texture textons or using color textons, but larger influence on that of using texture textons. Among the four metrics, the Euclidean distance has the highest accuracy for both color-texture textons and texture textons, whereas the Cityblock distance produces the highest accuracy for texture textons on both the cropped and image datasets.

f) For vegetation segmentation on both the cropped and image datasets, soil appears to be the most difficult one for correct segmentation and it is often confused with brown grass due to a similar yellow color particularly under a shining condition. A significant proportion of tree pixels are misclassified as road. Unlike the result on the cropped region data that there is little confusion between brown and green grasses, the discrimination between green and brown grasses is challenging in the natural images, probably due to the difficulty manually annotating pixelwise ground truths.

## 6. Conclusion and Future Works

This paper proposes a class-semantic texton based approach for vegetation segmentation in natural roadside images. It learns class-semantic color-texture textons for more effective representation of class specific features from training data, and then map features of all pixels into the learnt textons. A superpixel based collective classification strategy is used to label each superpixel by aggregating the combined occurrence of color-texture textons. We conduct experiments to investigate the optimal values of several key parameters of the proposed approach, which achieves the highest accuracies of 78.9% and 74.5% on two natural datasets and promising results on a set of real-world videos and the public Croatia roadside grass dataset. Our results indicate that shadows of objects and shining pose the biggest challenges for robust segmentation of vegetation, leading to overlap in the color between shadow and tree stem, as well as between brown grass and soil pixels. Furthermore, tree pixels are also prone to be misclassified as road. For accurate segmentation in natural conditions, it is desirable to consider features robust against lighting variations for the segmentation of these objects.

The proposed approach still can be extended in several aspects. Firstly, superpixel based classification employs all pixels within each superpixel without considering their label consistency with neighbouring superpixels. As a superpixel and its neighbours are highly likely to belong to the same category, it is advisable to incorporate collective decisions in larger superpixel neighbourhoods to enforce classification consistency on object categories. Aggregated features in superpixel neighbourhoods have exhibited good performance for object categorization [22]. Secondly, only pixel-level color and texture features are currently considered. As the characteristics of natural objects are often represented in a larger spatial region, it is worthy incorporating statistical features over regions [11] to generate more robust descriptors of objects. Thirdly, we will add global and local contextual features (e.g. object co-occurrence statistics) to further improve the performance of our approach. Our future work will also extend the natural dataset for a thorough evaluation of the proposed approach.

## Acknowledgements

This research was supported under ARC (Australian Research Council) Linkage Projects funding scheme (project number LP140100939).